\documentclass[sigconf]{acmart}

\usepackage[utf8]{inputenc}
\usepackage[T1]{fontenc}
\usepackage{hyperref}
\usepackage{url}
\usepackage{booktabs}
\usepackage{amsfonts}
\usepackage{nicefrac}
\usepackage{microtype}
\usepackage{subcaption}
\usepackage{amsmath}
\usepackage{mathtools}
\usepackage{algorithm}
\usepackage{xcolor}

\AtBeginDocument{
  \providecommand\BibTeX{{
    \normalfont B\kern-0.5em{\scshape i\kern-0.25em b}\kern-0.8em\TeX}}}

\copyrightyear{2021} 
\acmYear{2021} 
\setcopyright{acmcopyright}\acmConference[ICAIF'21]{2nd ACM International Conference on AI in Finance}{November 3--5, 2021}{Virtual Event, USA}
\acmBooktitle{2nd ACM International Conference on AI in Finance (ICAIF'21), November 3--5, 2021, Virtual Event, USA}
\acmPrice{15.00}
\acmDOI{10.1145/3490354.3494394}
\acmISBN{978-1-4503-9148-1/21/11}

\begin{document}

\title[Asynchronous Collaborative Learning Across Data Silos]{Asynchronous Collaborative Learning Across Data Silos}

\author{Tiffany Tuor}
\affiliation{
  \institution{J.P. Morgan AI Research}
  \city{London}
  \country{United Kingdom}
}
\author{Joshua Lockhart}
\affiliation{
  \institution{J.P. Morgan AI Research}
  \city{London}
  \country{United Kingdom}
}
\email{joshua.lockhart@jpmorgan.com}

\author{Daniele Magazzeni}
\affiliation{
  \institution{J.P. Morgan AI Research}
  \city{London}
  \country{United Kingdom}
}

\renewcommand{\shortauthors}{Tuor, Lockhart, Magazzeni.}

\begin{abstract}
Machine learning algorithms can perform well when trained on large datasets. While large organisations often have considerable data assets, it can be difficult for these assets to be unified in a manner that makes training possible. Data is very often `siloed' in different parts of the organisation, with little to no access between silos. This fragmentation of data assets is especially prevalent in heavily regulated industries like financial services or healthcare. In this paper we propose a framework to enable asynchronous collaborative training of machine learning models across data silos. This allows data science teams to collaboratively train a machine learning model, without sharing data with one another. Our proposed approach enhances conventional federated learning techniques to make them suitable for this asynchronous training in this intra-organisation, cross-silo setting. We validate our proposed approach via extensive experiments.

\end{abstract}
\maketitle
\section{Introduction}
\label{introduction}

Machine learning is experiencing a renaissance, spurred on by the exciting achievements of deep neural networks \cite{alphastar},
\cite{alphafold},
\cite{radford2021learning},
 \cite{devlin2018bert}. Machine learning algorithms can be applied to a wide variety of tasks relevant to the financial sector: to predict future events, to classify or cluster documents,
to segment customers or predict their behaviour, to detect fraudulent or anomalous transactions, to obtain useful insights from large amounts of data, \emph{etc.}
One key enabler of machine learning is access to large training data sets. The traditional view of training a machine learning model is to assume that all training data is available for access at one location. In practice, this is rarely true, especially for organisations in the financial services sector. Regulatory restrictions, privacy concerns, or simply mundane infrastructural challenges often impede the flow of data across the organisation. Data exists in, and can be divided across, a number of \textit{silos} within the organisation. Teams of engineers have access to some subset of these data silos, and develop machine learning models trained on data from those silos. Teams do not have a view of data that sits in another team's silo. Indeed, in larger organisations, they may not even know of the existence of other teams that work on similar problems. One solution to this fragmentation is to build an organisation-wide \textit{data lake}, a one-stop-shop from which data science teams can extract data for training. Depending on the size of the firm, its geographic footprint, the regulatory environment it operates in \emph{etc.}, such a data unification process can range from a major engineering effort to completely intractable.

\begin{figure}[h]
\includegraphics[width=8cm]{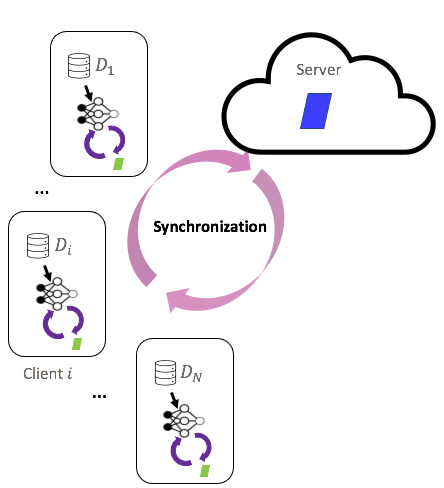}
\caption{Federated learning, schematic overview. Each client trains their own model on their own data, and communicates the weights to the server. The server combines the weights, and sends this back to the clients.}
\label{fig:overviewfl}
\end{figure}

In this paper we consider a middle ground between the silo approach and the firm-wide data lake by employing the \textit{Federated Learning} paradigm. 
Federated learning enables a group of distributed entities (\textit{clients}) to collaboratively train a machine learning model, without sharing their data with one another -- local data stays local~\cite{mcmahan2016communication}. Typically, a client trains their own \textit{local model} on their own data before sending this local model to a central server. This central server aggregates the models it receives from all clients into a \textit{global model}, then sends this global model back to each client. This train, aggregate, redistribute process continues iteratively until suitable accuracy has been attained (see Figure \ref{fig:overviewfl}).

The focus of this work is to tackle the problem of \textit{collaborative learning across data silos}. We consider the case of multiple data science teams who would like to collaborate on training a model for some prediction task. For whatever reason, these teams are not permitted to unify their respective datasets. The federated learning paradigm described above is ideal for our purposes, because no data is transferred between the clients. Furthermore, we are able to show that clients benefit from the collaboration in the form of increased classification accuracy for their individual tasks.
By enabling teams in different silos to train a model together, federated learning allows the teams to train models on a more diverse set of data samples, which consequently boosts model generalization and performance.

\begin{figure}[h]
\includegraphics[width=8cm]{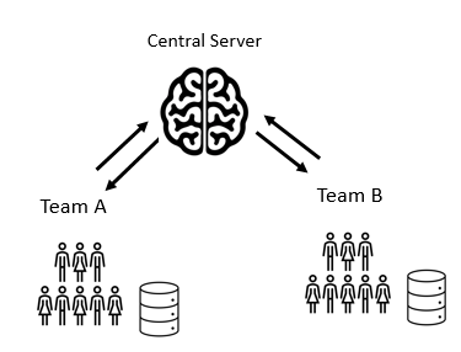}
\caption{Cross-Silo Federated Learning, with teams within an organisation.}
\label{Cross-Silo}
\end{figure}

One can distinguish two types of federated learning (FL) based on the \textit{number of clients} involved in the collaboration, and the \textit{size of their datasets} \cite{kairouz2019advances}. When the number of clients is very large, and the dataset size per client is relatively small, it is referred to as \textit{Cross-Device} FL. Use cases for Cross-Device FL include training models across many mobile devices, Internet-of-Things (IoT) gateways, sensors \textit{etc.} The second type of federated learning is often referred to as \textit{Cross-Silo} FL and is used when the number of nodes is small and the datasets are large. This is precisely the silo setting we described earlier, with multiple clients each with their own datasets stored locally. These clients wish to collaborate in training a model (see Figure \ref{Cross-Silo}).

Traditional federated learning \cite{mcmahan2016communication} assumes clients train simultaneously. In practice, this is rarely the case, as clients are rarely available at the same time. In Cross-Device FL, active devices may drop out of the training before the completion of the learning task due to connectivity or energy constraints. This issue has seen less attention within Cross-Silo FL, but becomes very relevant for our purpose of developing a scheme for collaborative training within an organisation. It is not a reasonable assumption that data science teams have the same training schedules, \textit{e.g.}, some teams may train monthly, others weekly. Even if the training cadence matches, the specific training times will often never overlap.

Clients that train in a non-overlapping fashion can have a major impact on the global model, and we show this later on in the paper. If the data held by the clients leaving the protocol is very different from others, the knowledge from these ``missing'' clients will be forgotten over time due to the \textit{catastrophic forgetting} phenomenon \cite{de2019continual}~(\textit{i.e.}, the global model will be updated without taking into account parameters from missing clients). In other words, as clients drop out or join, data available to train the global model varies over time (\textit{i.e.}, non-stationary), which causes the global model to forget previously acquired knowledge when learning new information.
\paragraph{Summary of contribution} 
\begin{itemize}
\item We show that federated learning can be adapted to enable \textit{collaborative learning across data silos}.
\item We show that teams benefit by collaborating via federated learning: (i) classification accuracy improves if another team collaborates with you, (ii) classification accuracy improves if you join a collaboration.
\item We show that synchronous federated learning techniques perform poorly at this collaboration when a teams have different training schedules. We present a technique for doing asynchronous federated learning in this setting by means of a \textit{coreset} to serve as a proxy for disconnected clients. We show that this technique performs well.
    
\end{itemize}

\section{Synchronous Federated Learning: Basic Concepts}
\label{section:federated}

In machine learning, a \textit{loss function} is used to evaluate how well a model fits its training data.
The loss function can be defined on a per-sample basis for a given model's parameters $\theta$, an input data sample $x_j$, and possibly (if it is a supervised model) a desired output label $y_j$. The overall loss function $L(\theta)$ is defined as the average of all per-sample loss functions for each sample in the training dataset.

The goal of model training is to find the parameters $\theta$ that minimize the loss function $L(\theta)$, which is equivalent to finding the best parameter $\theta$ that fits the training data $\mathcal{D}$. In what follows we refer to the model specified by the parameters $\theta$ as the global model. Usually, an optimizer such as gradient descent is used to minimize the loss function.

In the federated learning setting, each of the $N$ clients has a `local' loss function $L_i({\theta})$ defined on its local training dataset $\mathcal{D}_i$. The global loss function of all clients is then defined as:
\begin{equation}
\label{eq:global-loss}
    L(\theta)=\frac{\sum_{i=1}^{N} |\mathcal{D}_i|L_i(\theta)}{\sum_{i=1}^{N} |\mathcal{D}_i|},
\end{equation}
where $\left|\mathcal{D}_i\right|$ denotes the number of data points in the training set $\mathcal{D}_i$.
The basic federated learning approach to the minimization of  the global loss function $L(\theta)$ is to apply \textit{distributed gradient descent} \cite{mcmahan2016communication}:
\begin{enumerate}
    \item Each client (simultaneously) minimizes its local loss function $L_i(\theta)$ by performing $\tau$ steps of gradient descent on its local model parameter $\theta_i$, according to
    \begin{equation}
    \label{local}
    \theta_i(t)= \theta_i\left(t-1\right)-\eta \nabla L_i\left(\theta_i\left(t-1\right)\right),
    \end{equation}
    where $\eta$ is the learning rate.
    \item  Each client sends its resulting parameter $\theta_i(t)$ to the server.
    
    \item The server aggregates the parameters received from each client, according to
    \begin{equation}
    \label{eq:global-agg}
    \theta(t)= \frac{\sum_{i=1}^{N} \left|\mathcal{D}_i\right|\theta_i \left(t\right)}{\sum_{i=1}^{N} \left|\mathcal{D}_i\right|} .
    \end{equation}
    
    \item The server sends the global model parameter $\theta(t)$ computed in (\ref{eq:global-agg}) to each client. After receiving the global model parameter, each client updates its local model parameter $\theta_i(t) \leftarrow \theta(t)$.
     
\end{enumerate}
 
The protocol described above is synchronous, as it assumes the server receives updates from all clients before performing parameter aggregation (\textit{i.e.}, Equation (\ref{eq:global-agg})). We refer to this as the Synchronous federated learning setting. In practice however, the server must be capable of acting asynchronously due to client heterogeneity: a client that has access to less compute power will slow the above process down; also it is very common for clients to drop out of the training process completely, which would cause the above protocol to stall indefinitely.
Thus, synchronous federated learning is rarely feasible in practical scenarios. 

Let us consider practical issues with our collaborative training setting, where multiple data science teams train without sharing data between silos. Requiring all teams to adopt the same training schedule is unrealistic. Some teams will train monthly, others weekly, some perhaps on an ad-hoc basis. Furthermore, the requirement that their compute infrastructure is ready to train at any given time just so they can participate in the collaboration is also unrealistic, as it assumes their infrastructure is available on demand. In the next section we outline how federated learning can be performed asynchronously.

\section{Asynchronous Federated Learning and Continual Learning}
 
In order to mitigate the above-outlined practical issues around synchronous federated learning, a naive strategy would be for the central server to aggregate only the weights of those clients that are available at each particular moment. This would mean that at time step $t$, the server would aggregate the local parameters received from the subset of clients who happen to be connected at time $t$, indexed $K_t\subseteq \{1,\dots,N\}$. Hence, Equation (\ref{eq:global-agg}) in Section \ref{section:federated} would become:
\begin{equation}
        \label{eq:global-agg2}
    \theta(t)= \frac{\sum_{i\in K_t} |\mathcal{D}_i|\theta_i (t)}{\sum_{i \in K_t} |\mathcal{D}_i|} .
\end{equation}
where the subset $K_t$ changes over time according to client availability.

\begin{figure}[ht] 
  \begin{subfigure}[b]{1\linewidth}
    \centering
    \includegraphics[width=6cm]{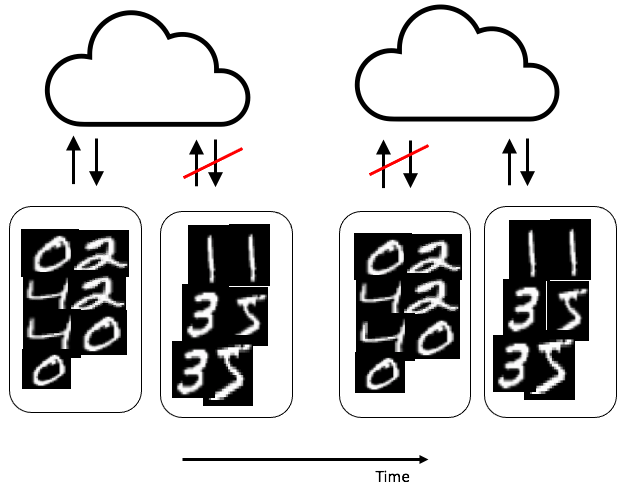}
    \caption{Data across clients is non-i.i.d., note that the dataset at each time step is different.} 
    \label{1a} 
    \vspace{4ex}
  \end{subfigure}
  
  \begin{subfigure}[b]{1\linewidth}
    \centering
    \includegraphics[width=6cm]{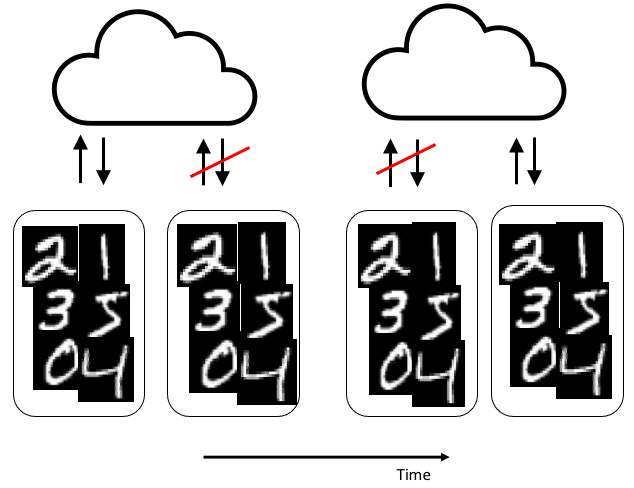}
    \caption{Data across client is i.i.d., note that the dataset at each time step is the same.} 
    \label{1b} 
    \vspace{4ex}
  \end{subfigure}
  \caption{Different data distributions across clients: time proceeds from left to right, red lines denote disconnection of device from cloud.} 
 \end{figure}

If the union of all client data is independent and identically distributed (i.i.d.), as illustrated in Figure (\ref{1a}), then using Equation (\ref{eq:global-agg2}) as a substitute of Equation (\ref{eq:global-agg}) has only a minor impact on the learning process. In this i.i.d. setting, the classification problem is stationary, regardless of the subset of the clients that are connected at any given time. However, when we are not guaranteed data that is distributed identically as client index changes, that is, when it is possible that some client's dataset is drawn from a different distribution than the other clients (see Figure \ref{1b}), then the training data for the global model is \textit{non-stationary}.
This means that using Equation (\ref{eq:global-agg2}) as substitute of Equation (\ref{eq:global-agg}) will result in the erasure of knowledge acquired from the disconnected clients. This phenomenon is known as \textit{catastrophic forgetting} \cite{de2019continual}. That is, if training data arrives in an online and non-stationary fashion then the global model can fail to consolidate previously attined knowledge: the old data is forgotten.

In the following, we will address the catastrophic forgetting problem  that occurs in asynchronous federated learning. Our goal is to allow the global model to acquire new knowledge from new subsets of clients while preventing the new input from significantly interfering with existing knowledge.

\section{Problem Formulation}
\label{sec:strategy1}
We assume that there are $N$ clients. We assume that over a fixed, discrete time experimental period these clients can join or leave the collaborative training
whenever they want. For example, a client joins at time $t=1$ and begins training. Another client joins at time $t=3$. The first client leaves at time $t=4$, the second client chooses to leave at $t=6$.

The goal is to learn a global model $\theta_T$ which has high accuracy on the union of the clients' datasets $\mathcal{D}_1 \cup \mathcal{D}_2 \cup \dots \cup \mathcal{D}_N$. This is challenging because the subset of the clients that are connected changes throughout training. If the global model was trained solely on the dataset of available clients, the knowledge acquired from disconnected clients will be erased. 

To summarise, our goal is to find an efficient \textit{federated continual learning} approach that:
\begin{itemize}
    \item has the ability to train a global model on non-stationary data (\textit{i.e.}, different subset of clients connected over time with non-i.i.d. data distribution \textit{across} clients)
    \item provide at any time a competitive global model with satisfactory classification accuracy for all the data observed so far (\textit{i.e.}, all the data available at the clients that have been connected to the collaboration at least once),
    \item have computational and memory requirement bounded or growing very slowly with respect to the data already seen.
\end{itemize}

\section{Proposed Approach}
We propose a heuristic approach to federated learning that is able to learn a competitive model with good accuracy on the union of client's datasets, despite clients training asynchronously. 

\subsection{Proxy Dataset and Proxy Model}
During their first connection to the collaboration/training, each client provides the server with a small representative summary, $\mathcal{C}_i$, of its local data set. Such a summary of a dataset is often referred to as a \textit{coreset}~\cite{lu2020sharing} or proxy dataset. This coreset is usually much smaller than $\mathcal{D}_i$ and can act as proxy of the client $i$ when it is disconnected. A multitude of techniques exist to construct a coreset from a dataset \cite{lu2020sharing,feldman2011unified}.
Thus, the central server can simulate the missing clients by training a local model, $\hat\theta_i$ for all missing clients using their proxy datasets.

\subsection{Training with proxy datasets}
During training, the server always performs aggregation with as many model updates as total number of clients (\textit{i.e.}, $N$), however in the cases where a client is disconnected, model updates are replaced by proxy model updates, as follows:
  \begin{equation}
    \label{eq:global-agg-async}
    \hat\theta(t)= \frac{\sum_{i=1}^{N} |\mathcal{D}_i|\tilde\theta_i (t)}{\sum_{i=1}^{N} |\mathcal{D}_i|} .
\end{equation}

where  $\tilde \theta_i(t) = \theta_i(t)$ (see Equation (\ref{local})) if client $i$ has provided an update at time step $t$. If no update was provided by the client, a proxy model is used instead, $\tilde \theta_i(t) = \hat\theta_i(t)$. The server trains the proxy model by performing $\tau'$ steps of gradient descent on the coreset ${C}_i$:
\begin{equation}
    \hat\theta_i(t) = \tilde \theta_i(t-1) -\eta \hat\nabla L_i(\tilde\theta_i(t-1))
\end{equation}

where $\hat\nabla L_i$ is computed only on the coreset. Note that the number of local updates performs by the server when training the proxy model (\textit{i.e.}, $\tau'$) differs from the number of local updates performs by the clients ($\tau$). 

It is important to give some thought to privacy aspects of our protocol. The privacy requirements of our intra-organisation collaborative learning scenario is that there is no data leakage between data silos. We assume that there is no issue with teams sending data to a central entity for training, since that central entity is owned by the firm. The fact we store a coreset on that central entity, while anathema in standard privacy-focused FL settings, is reasonable for us because that central entity is trusted. To protect against external data leaks from that central entity, all coresets can of course be encrypted and decrypted when needed. Future work should focus on this aspect: our coreset-based protocol could possibly be made secure by means of secure multiparty computation protocols \cite{smpc}.

\section{Related Work}
Federated learning is a term coined by McMahan et al. in \cite{mcmahan2016communication} which refers to a set of techniques that allow several entities to collaboratively train a machine learning model, in a distributed manner. Surveyed in \cite{kairouz2019advances} and \cite{Li2020FederatedLC}, these techniques consider the setting where training is carried out locally by each entity involved in the protocol. All of these local models are periodically combined in some way, with the aim of improving overall performance. The key idea is that no data is shared between entities taking part in the protocol.

A rich source of applications of federated learning comes from mobile computing \cite{lim2020federated}, \emph{e.g.}, the predictive keyboard functionality in smart phones is an informative example \cite{hard2019federated}, \cite{yang2018applied}: while it makes sense for all users of the app to contribute to training of a global predictive model, it is important from a privacy perspective that what a user types on their phone is not shared with a central server, or with other users of the keyboard app.
See also uses in open banking \cite{long2020federated}.
The former setting is referred to as cross-device federated learning: working with large numbers of devices, often with low bandwidth and computational power. Another setting for federated learning, and more relevant for this work, is the \textit{cross-silo} setting, categorized by Kairouz et al.~\cite{kairouz2019advances} as having fewer clients (of the order of 10s to 100s), that are always connected to the central entity for training. Concretely, consider a network of hospitals who wish to collaborate in training a predictive model to diagnose patients, but who cannot share confidential patient data with one another \cite{huang2020loadaboost}, \cite{sheller2018multiinstitutional}. Other settings include the agri-food sector \cite{durrant2021role}, and detection of financial crime \cite{suzumura2019federated}.

Federated learning in an asynchronous setting is a challenging proposition, but necessary if we consider the realities of our collaborative training scenario. Certainly in the cross-device setting it is not reasonable to assume that all devices are heterogeneous in terms of computational power and bandwidth \cite{Sahu2020FederatedOI}, \cite{McMahan2017CommunicationEfficientLO}, \cite{Konecn2016FederatedLS}. 

Due to this heterogeneity, one client often slows down the whole process (referred to as the \textit{straggler effect}) as in the  synchronous setting the server needs to wait for model updates from all clients before proceeding to the next round. To mitigate this problem, some techniques for asynchronous FL have been proposed \cite{xie2020asynchronous},\cite{van2020asynchronous},\cite{chai2020fedat} which enable the server to proceed with training only those model updates that are currently available (despite some updates being potentially stale) and consequently to avoid the need to wait for all clients. Some re-weighting schemes are employed which take into account the degree of staleness of the weights. However, these techniques are reasonable if the delay is relatively \textit{small}. In the case where clients drop out of the collaboration, the delay can become very large and the existing approaches are no longer suitable. Note that these approaches tend to target Cross-Device FL settings, where the fact that there are many devices involved in the collaboration increases the probability of a large number of devices being connected at a given time. Naturally, with Cross-Silo FL the number of clients is much smaller, which means there can be periods of time where very few clients are overlapping in their training. This setting is typical in our intra-organisation setting, where the model training regimen of each team is rarely the same.

\section{Experiments}
\label{sec:work}
We are interested in the notion of cross-team collaboration within a single organisation, and so focus on conducting experiments that test desirable properties of a system that would enable such collaboration. 
\subsection{Datasets and testing details}
\label{section:datasets_and_test_details}
In what follows we consider an experimental test-bed containing two clients, with the crucial issue that either client can leave the collaboration at any time. We emphasise the following desiderata for our system:
\begin{enumerate}
\item It should be beneficial for clients to collaborate with one another: clients should be tempted to join and stay in the collaboration by the promise of higher performance than if they had not joined;
    \item performance should not be impacted by a client leaving the collaboration while it is ongoing.
\end{enumerate}

Our experiments are conducted on the following datasets
\begin{itemize}
    \item \textbf{MNIST-SVHN} \citep{shin2017continual} is a split task dataset for two clients. The first client has the MNIST handwritten digits dataset, the second client has the Street View House Numbers (SVHN) dataset \citep{netzer2011reading}. The MNIST dataset has 60,000 training data samples, and the SVHN dataset has 73,000 training data samples.
    \item \textbf{Rotated MNIST} \citep{lopez2017gradient} is the split task dataset where each client has data from the MNIST handwritten digits dataset that have all been rotated by a fixed angle. In our experiments, we have two clients, the first with digits rotated by $0^{\circ}$ angle, the second with digits rotated by $90^{\circ}$. Each client has 60,000 training data samples. 
    \item \textbf{20NG1-20NG2} consists of a split task defined on the \textbf{20-Newsgroups} dataset \cite{newsgroup}. This dataset contains text posts to 20 different newsgroups, each on a different topic (e.g., Sport, Tech, Politics, Religion). In our experiments, the first client has data from the first 10 classes, the second client has data from the last 10 classes. Each client has approximately 8,000 training samples. We refer to each client's dataset as 20NG1 and 20NG2. 
\end{itemize}
We chose these datasets specifically to test the performance of our technique on a variety of data modalities (images, text), but also on a variety of model architectures (MLP, CNN). We note that performing well at document classification, specifically email classification (as represented in 20-Newsgroups) is very important in financial services. 

\subsection{Model architecture and training details}
As we noted in the previous section, we explicitly chose a variety of different data media for our experiments in order to ensure our techniques work with different model architectures. Let us now explicitly describe the model architectures and training procedures we used for these experiments.

For MNIST-SVHN, we use a convolutional neural network\footnote{The CNN has $7$ layers with the following structure: $5 \times 5 \times 32$ Convolutional $\rightarrow$ $2 \times 2$ MaxPool $\rightarrow$ $5 \times 5 \times 32$ Convolutional $\rightarrow$ $2 \times 2$ MaxPool $\rightarrow$ $1568 \times 256$ Fully connected $\rightarrow$ $256 \times 10$ Fully connected $\rightarrow$ Softmax.}. For the Rotated MNIST and 20NG1-20NG2 data, a fully-connected network with one hidden layer is used, with 200 and 1500 neurons (on the hidden layer) respectively. 

For all experiments, we use a batch size of 200. The learning rate varies according to experiments. The number of weight updates between \textit{training slots} (see details in next section) in the collaboration also depends on the experiments. 
For our proposed approach, the coreset size is set to $5\%$ of the training dataset, unless stated otherwise. We run each experiment 5 times and consider mean and standard deviation of the corresponding performance metrics.

\subsection{Performance Evaluation}
To evaluate performance of the models, we report the client-wise accuracy, by which we mean the accuracy of the model evaluated on each client's own local test dataset. As can be seen from the figures, mean accuracy is recorded as being over training \textit{slots}. A slot is defined as a fixed set of $k$ model updates, where $k$ is defined appropriately for each experiment. Note that $k$ is not to be interpreted as a hyperparameter, indeed it has no impact on model performance. Rather, for each experiment there is an appropriate update cadence, and the accuracy needs to be recorded in a generic way over all experiments. Thus, we need some way of specifying how often that accuracy is recorded. For each slot, we also record the standard deviation of the accuracy metric.

\subsection{Baselines}
It is useful to discuss several baselines that we feel are appropriate for our setting of \textit{collaborative learning across data silos}. These baselines serve to provide us with lower bounds, and upper bounds, on the possible accuracy of our technique.
\begin{itemize}
\item The \textit{Standalone} baseline, where each client trains a model on their own local data. No collaboration between clients. This provides us with a lower bound on performance.
\item The \textit{Ideal} baseline, where all clients train collaboratively and simultaneously according to the Fed-Avg protocol \cite{mcmahan2016communication}. No clients disconnect. This is intended to provide us with an upper bound on the possible performance of our model.
\item The \textit{Fed-Avg}\cite{mcmahan2016communication} baseline, in the setting where clients disconnect. This is intended to show the performance of using vanilla federated learning techniques in our asynchronous setting.
\end{itemize}
Since our technique makes use of a coreset stored on the central server for clients that are not currently engaged in the training protocol, it is useful to consider the \textit{Ideal} baseline through this lens. Specifically, this baseline corresponds to a protocol where all client data is stored on the central server, and the server takes over a client's training if they disconnect. That is, this corresponds to an identical protocol to ours, but where each client contributes a coreset equal to their full training data set.
\subsection{Results}
\subsubsection{Quantifying the benefit of collaboration}
The first set of experiments we conduct aim to 
quantify the benefit to each team of collaborating (desideratum 1, from Section \ref{section:datasets_and_test_details}). We do this by comparing the \textit{Ideal} baseline to the \textit{Standalone} baseline. Recall that the \textit{Ideal} baseline is the synchronous setting, where all clients are always connected, and weights are combined using Fed-Avg. The \textit{Standalone} baseline, in contrast, is where each client trains on their own dataset and there is no collaboration with other clients. That is, client weights are not combined in any way.

\begin{figure}
  \begin{subfigure}[b]{0.5\columnwidth}
        \centering
        \includegraphics[width=1\linewidth]{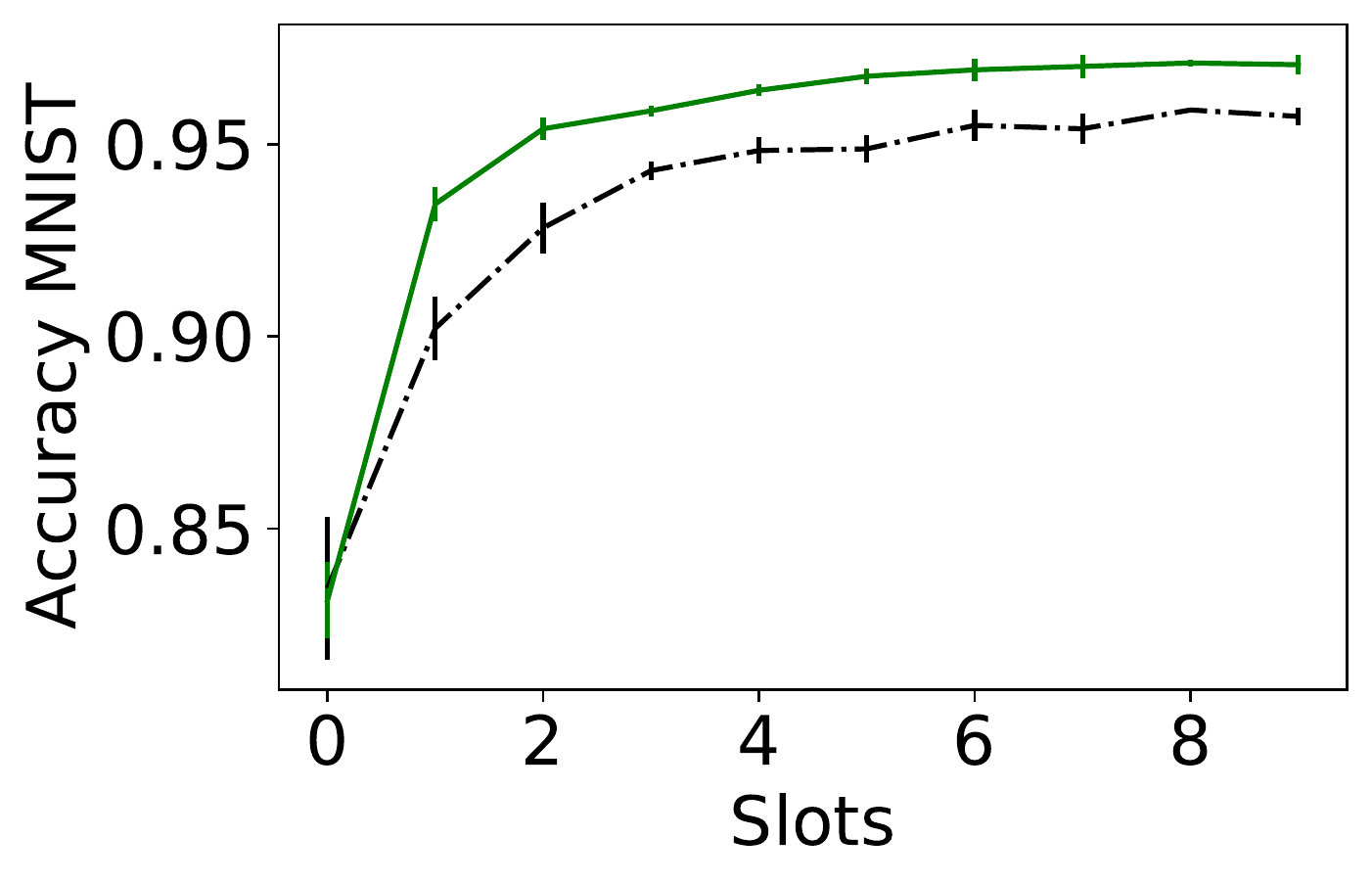}
    \end{subfigure}%
    ~
    \begin{subfigure}[b]{0.5\columnwidth}
        \centering
        \includegraphics[width=1\linewidth]{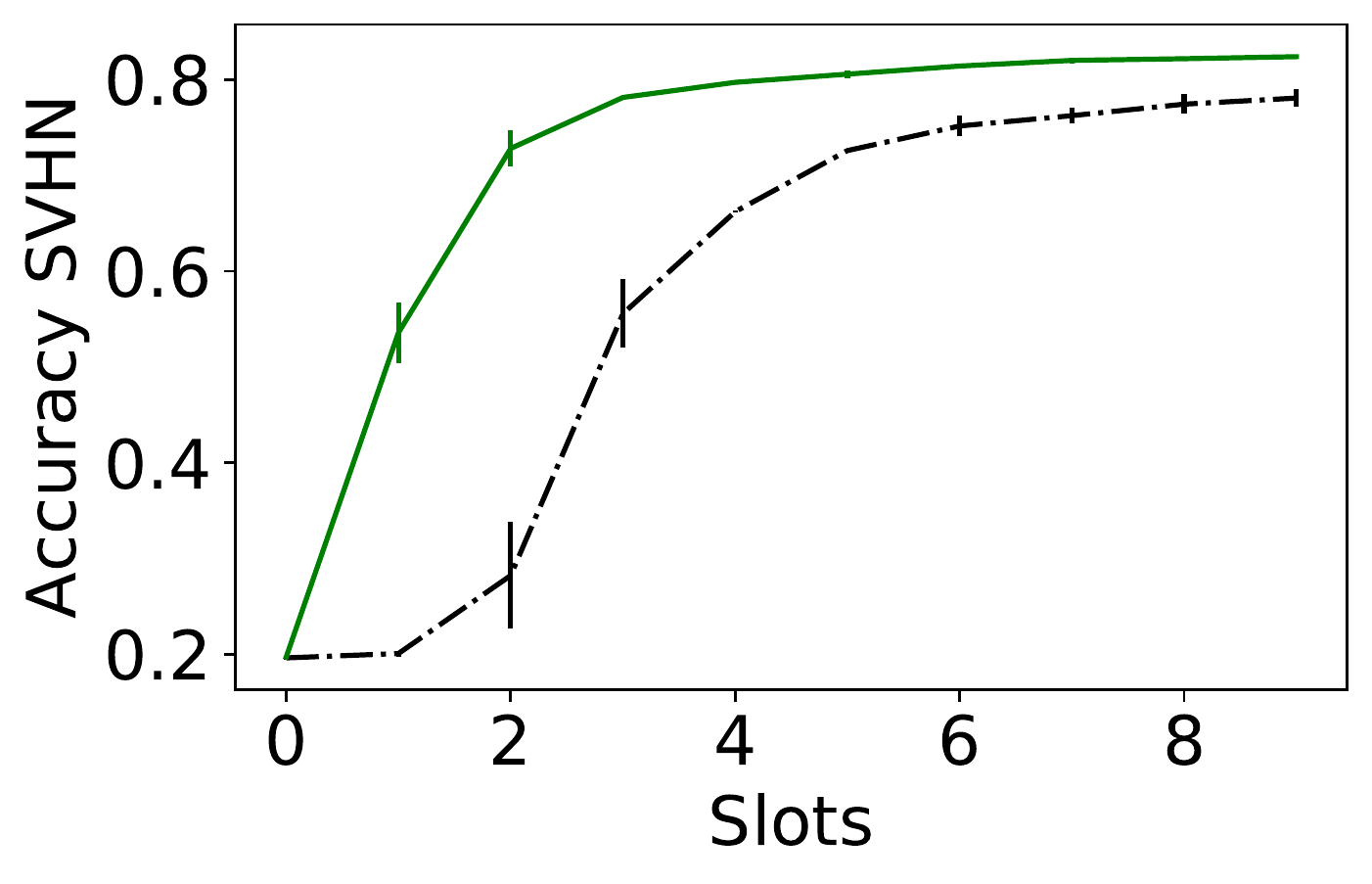}
    \end{subfigure}%
\caption{\textit{Ideal} (solid green line) VS \textit{Standalone} (dashed black line) for MNIST-SVHN}
\label{fig:1}
\end{figure}

\begin{figure}
    \begin{subfigure}[b]{0.5\columnwidth}
        \centering
        \includegraphics[width=1\linewidth]{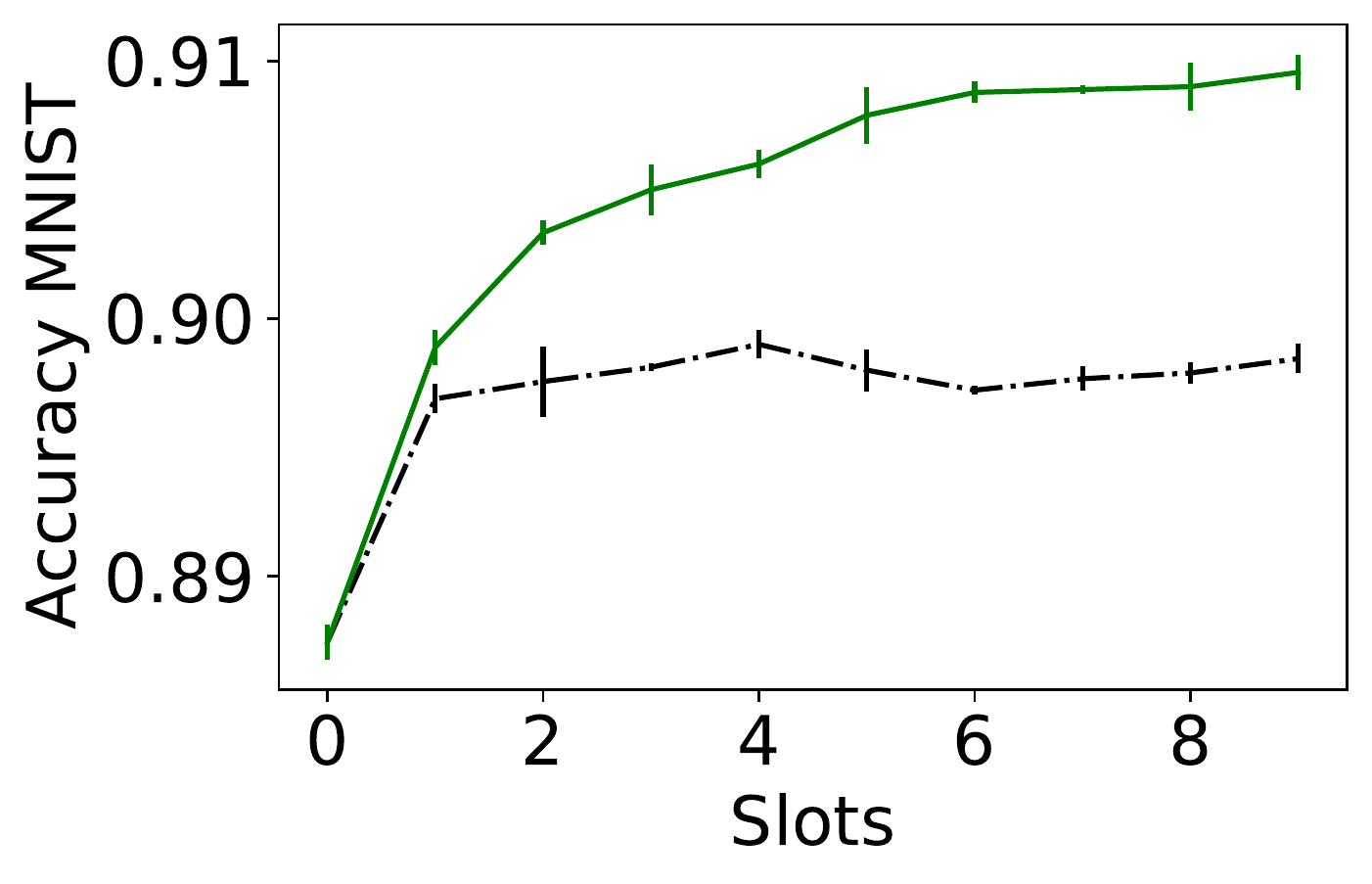}
    \end{subfigure}
        ~ 
    \begin{subfigure}[b]{0.5\columnwidth}
        \centering
        \includegraphics[width=1\linewidth]{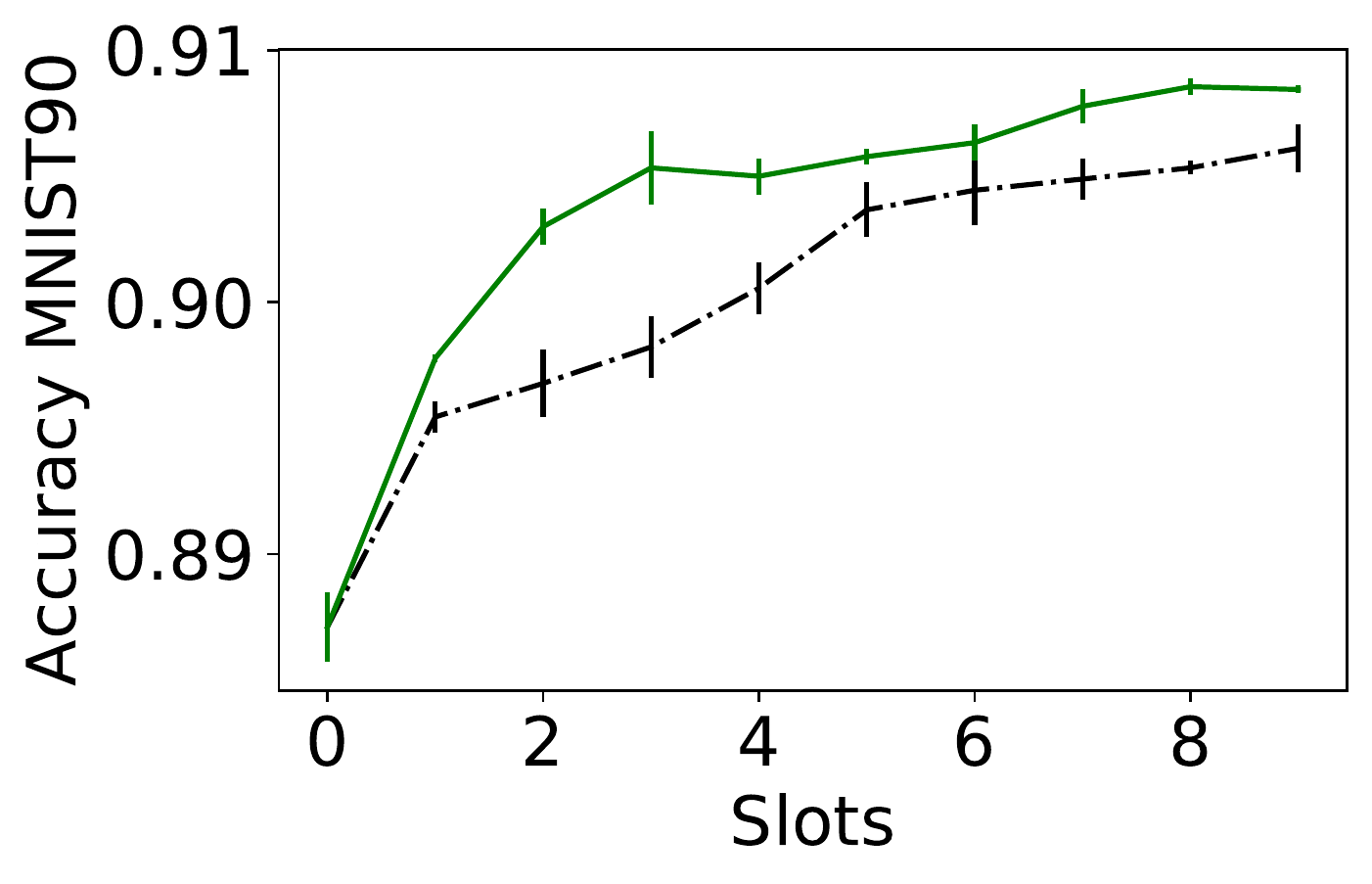}
    \end{subfigure}%
\label{exp:mnist}
\caption{\textit{Ideal} (solid green line) VS \textit{Standalone} (dashed black line) for MNIST-MNIST90}
\end{figure}

\begin{figure}
    \begin{subfigure}[b]{0.5\columnwidth}
        \centering
        \includegraphics[width=1\linewidth]{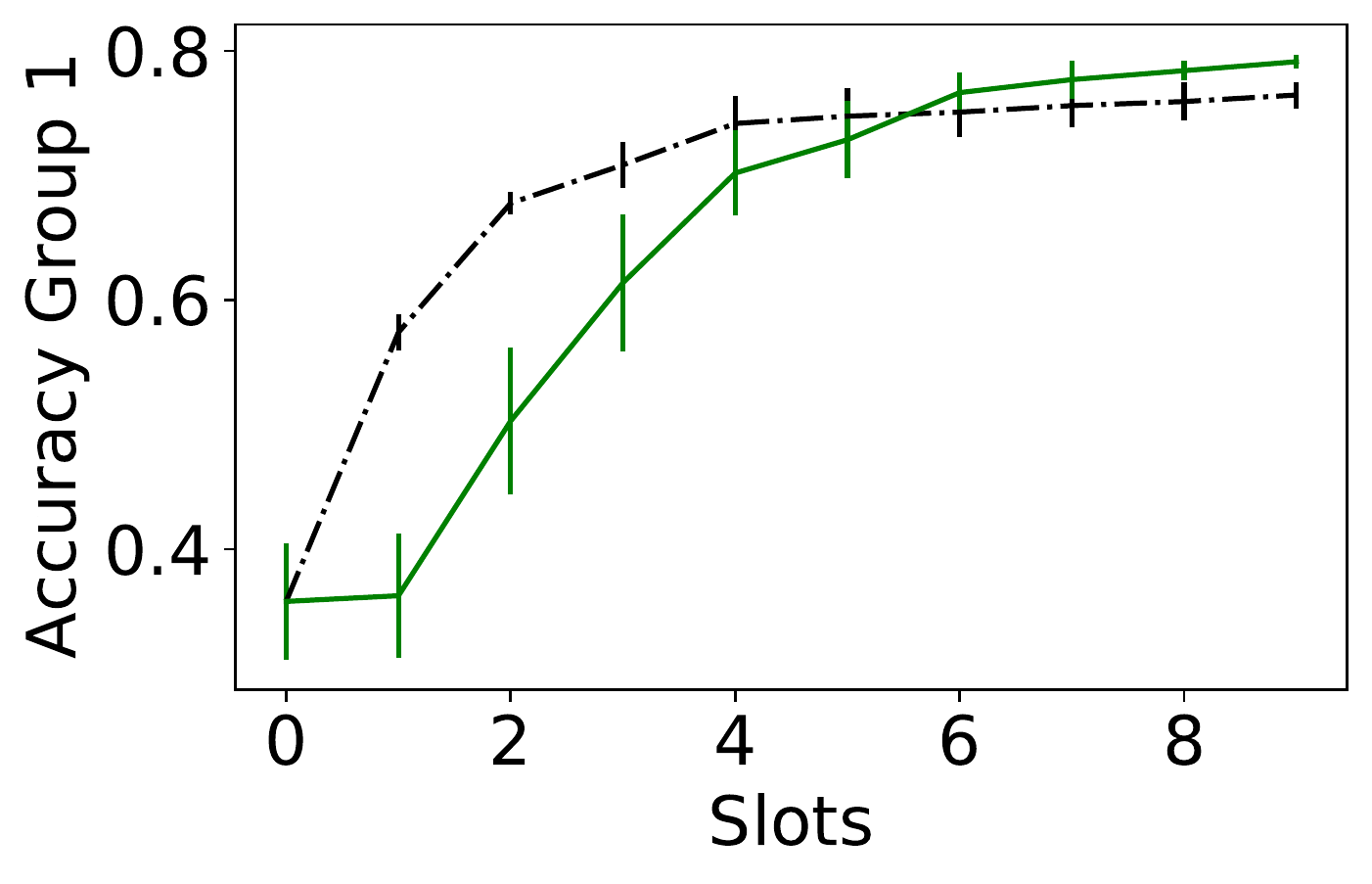}
        \label{e}
    \end{subfigure}%
    ~
    \begin{subfigure}[b]{0.5\columnwidth}
        \centering
        \includegraphics[width=1\linewidth]{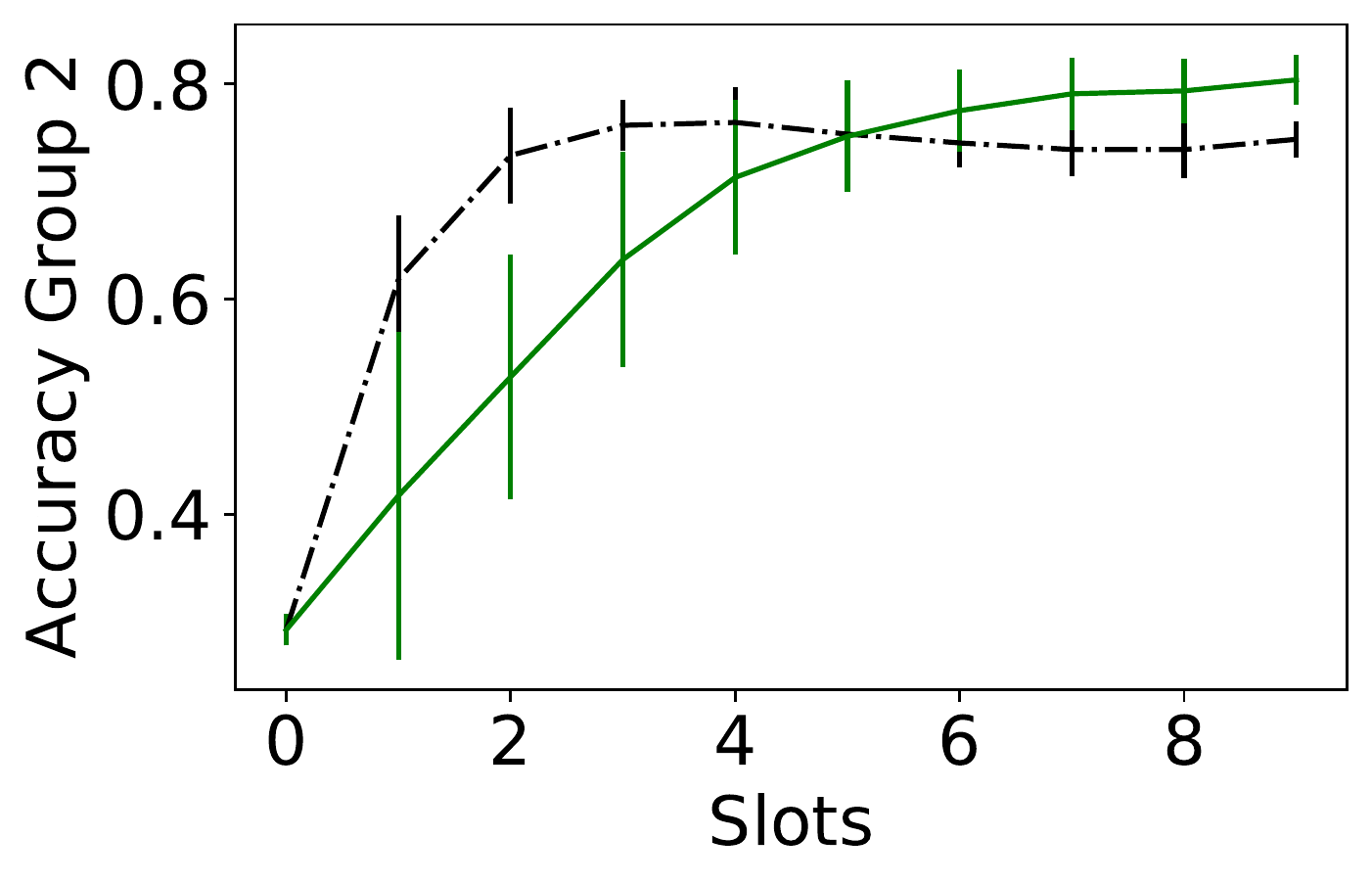}
        \label{f}
    \end{subfigure}%

    \caption{\textit{Ideal} (solid green line) VS \textit{Standalone} (dashed black line) for 20NG1-20NG2}
    \label{fig:3}
\end{figure}

Figures \ref{fig:1} through \ref{fig:3} show the results for MNIST-SVHN,  MNIST-MNIST90, and 20NG1-20NG2 respectively. For the majority of the experiments we observe that collaboration (solid green line) improves the performance of the local client models. In most cases, both clients benefit from the collaboration, but there are some cases where the benefit is not symmetric, that is, one client benefits more than the other. For example in Figure \ref{fig:1}, one can observe that the SVHN client benefits from an increase of $10\%$ in accuracy from client MNIST whereas the latter benefits only from a $3\%$ in accuracy from SVHN client. We venture that this is because MNIST is quite an easy task to perform well at, even without collaboration. 

\subsubsection{Measuring the performance of asynchronous training}
\begin{figure}
     \begin{subfigure}[b]{0.5\columnwidth}
        \centering
        \includegraphics[width=1\linewidth]{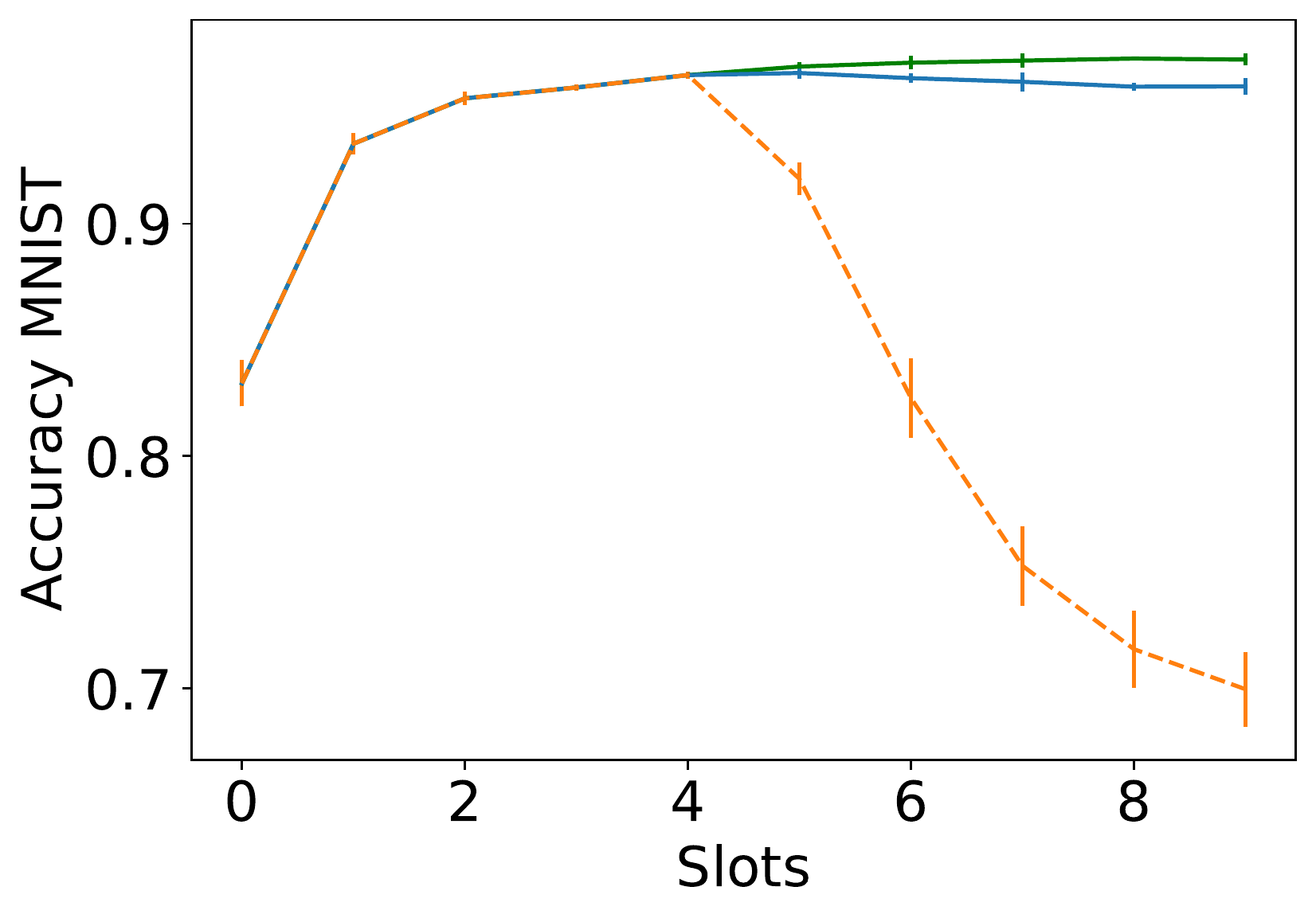}
        \caption{MNIST drops after slot 4.}
    \end{subfigure}%
    ~
    \begin{subfigure}[b]{0.5\columnwidth}
        \centering
        \includegraphics[width=1\linewidth]{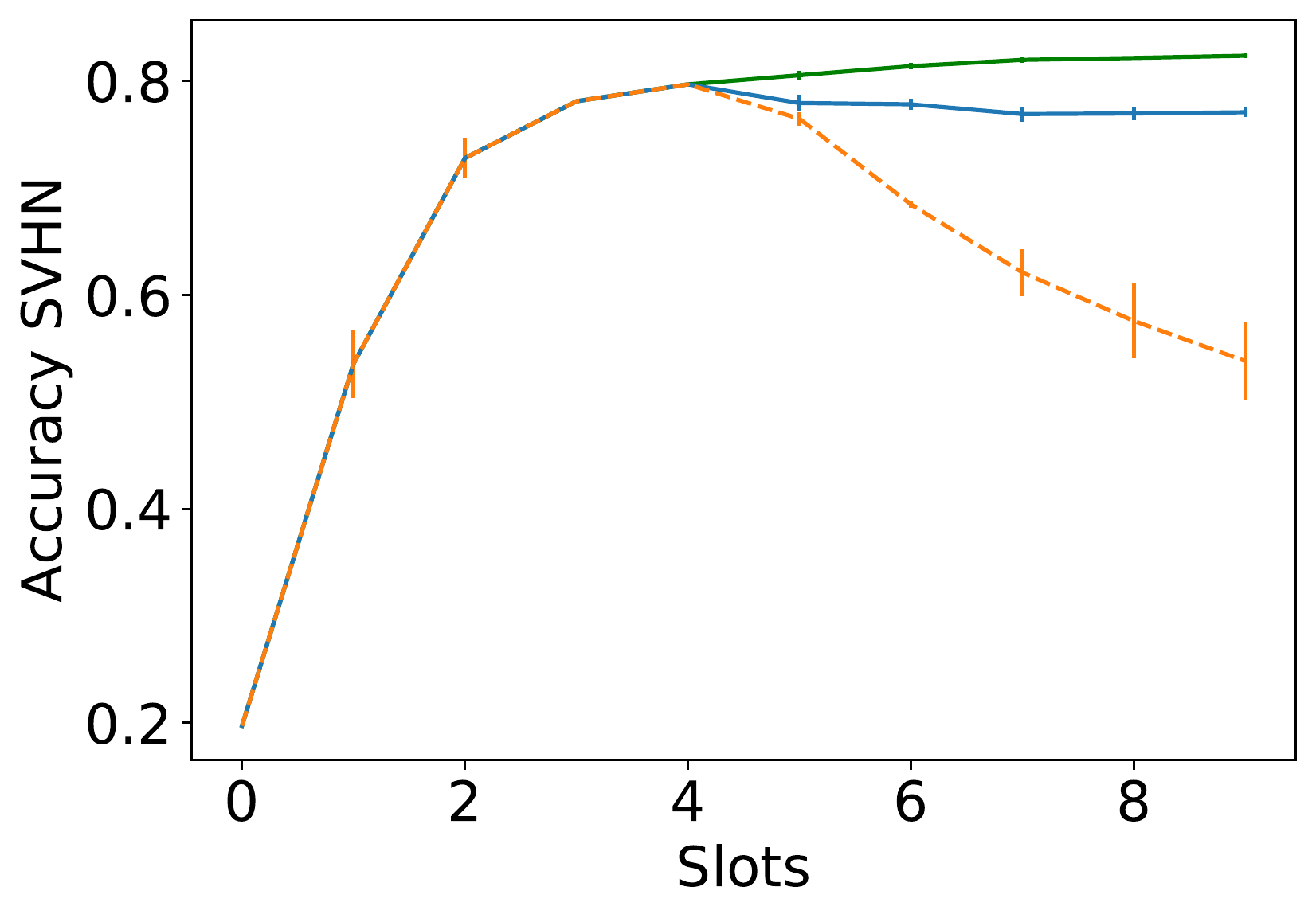}
         \caption{SVHN drops after slot 4.}
    \end{subfigure}%
 \caption{MNIST-SVHN : Asynchronous training, Ideal Baseline (green), FedAvg (orange), Proposed Approach (blue).}
\label{fig:dc-svhn}
\end{figure}

\begin{figure}    
    \begin{subfigure}[b]{0.5\columnwidth}
        \centering
        \includegraphics[width=1\linewidth]{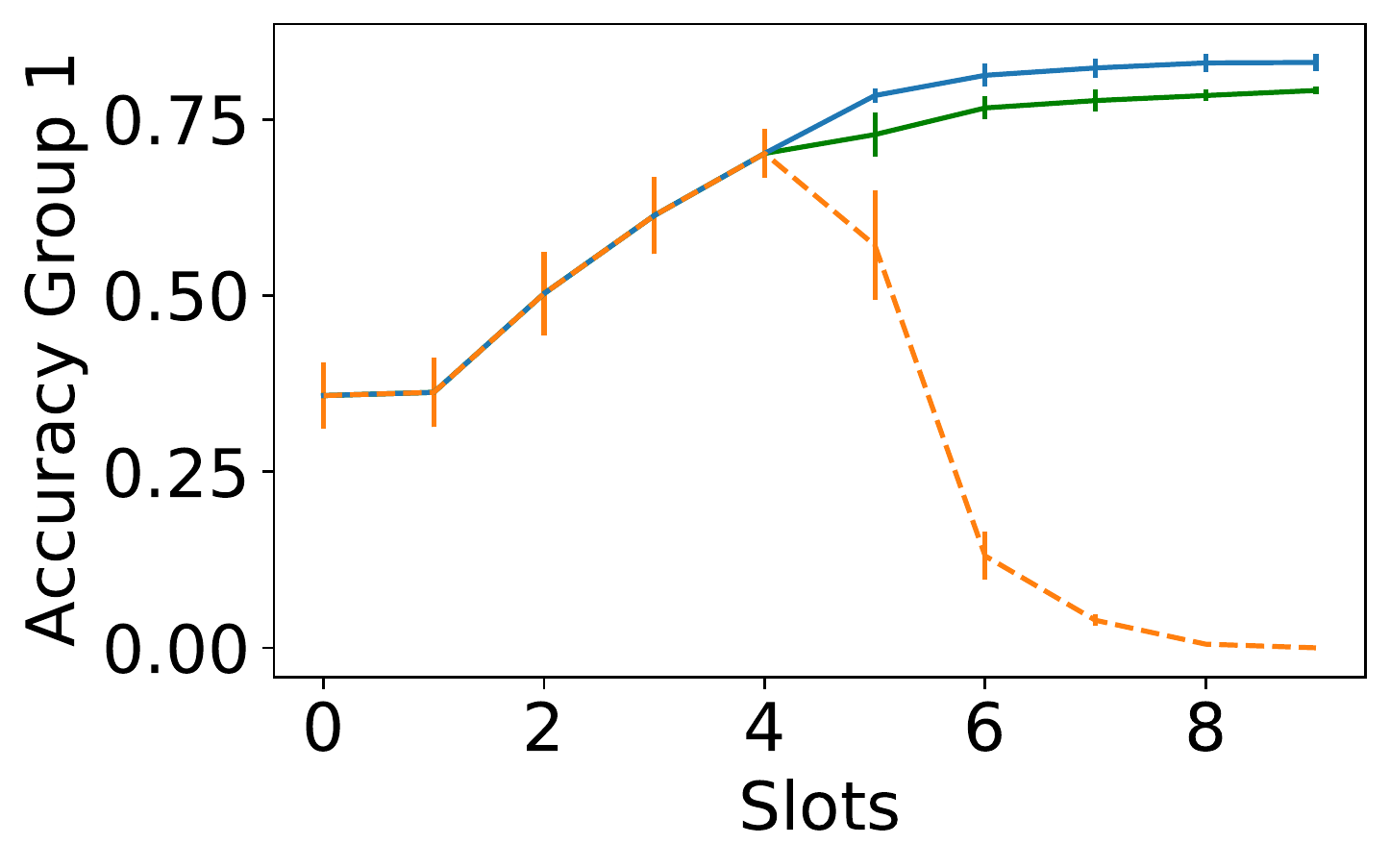}
        \caption{Group 1 drops after slot 4.}
    \end{subfigure}%
    ~
    \begin{subfigure}[b]{0.5\columnwidth}
        \centering
        \includegraphics[width=1\linewidth]{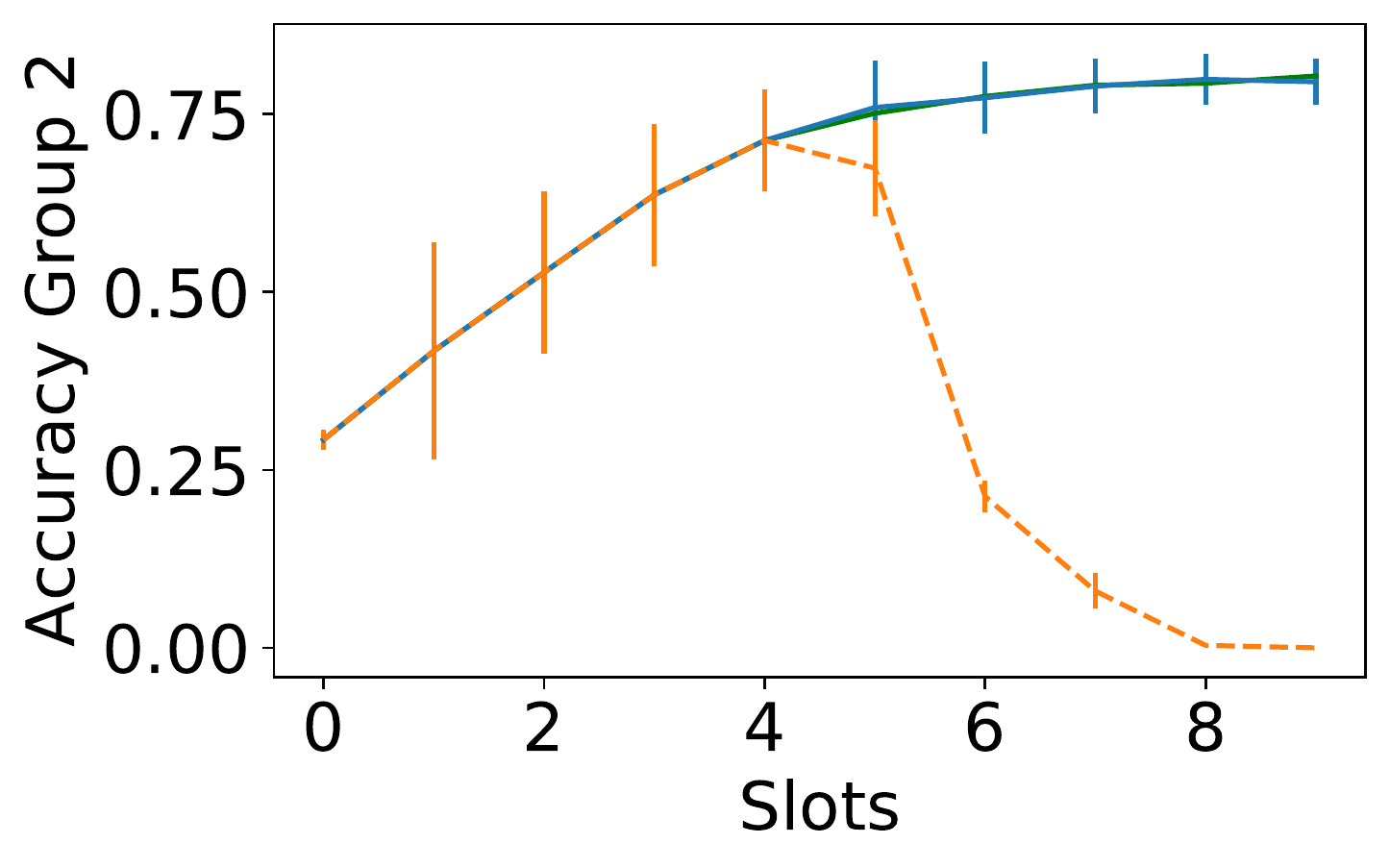}
        \caption{Group 2 drops after slot 4.}
    \end{subfigure}%
    \caption{20NG1-20NG2 : Asynchronous training, Ideal Baseline (green), FedAvg (orange), Proposed Approach (blue).}
    \label{fig:dc-group}
\end{figure}

Let us now consider clients that train in an asynchronous fashion. The first set of results (i.e., for datasets MNIST-SVHN, MNIST-MNIST90, and 20NG1-20NG2) are shown in Figures~\ref{fig:dc-svhn} through \ref{fig:dc-group}. In this setting, the two clients start by training together, but at slot 4 one of the clients stop contributing to the collaborative training. In each of these plots, the results where the first client drops are reported in plot (a), and where the second client drops in plot (b). On these plots we compare the performance of our proposed approach (Blue) with the \textit{Ideal} baseline (Green), and the \textit{FedAvg} baseline (Orange). We show the client-wise accuracy, which we take to be the accuracy of the collaborative model (that is, the model obtained according to that baseline's corresponding technique) evaluated on each client local test dataset. 
It is clear from these results that if nothing is done to handle the disconnection of a client then the model accuracy (that is, the accuracy of that client's model) decreases dramatically (see \textit{Fed-Avg} baseline, Orange line). We propose that this drop in performance accuracy occurs because once the client disconnects, they stop training and contributing to the global model. Then the global/collaborative model will forget the ``knowledge'' obtained from that client's data. On the other hand, when using our approach (Blue line), the global/collaborative model maintains a good performance on the data of the missing client even though that client is no longer contributing to the training. 
This is of course due to the fact that the central server is handling the training of the model on that missing client's coreset. This coreset is acting as a ``proxy'' for the missing client. Note that our proposed approach (Blue line) performs almost as well as the Ideal baseline (Green line) for all experiments.

\section{Open Challenges}
\label{section:open_challenges}
In completing this work we have uncovered several interesting challenges that can be addressed in future work.
The first of these is related to the privacy / security aspects of the protocol. Since we have focused on a specific, financial services appropriate setting, namely that of cross-silo collaborative training within a single organisation, there is no real need to consider the case where teams within that organisation are reluctant to share a data sample with some trusted central entity. However, if we were to extend this setting in the natural direction, that is, to allow the potential for collaborators outside the firm to be involved in the training, then security of the stored coresets becomes a pressing issue. Reiterating a point made in an earlier setting, we consider this protocol to be a potential use case for homomorphic encryption schemes.

The second challenge is related to the question of how to measure the contribution of each client involved in the collaborative training.
As we have seen in the experiments, the benefit received from the collaboration in terms of increase in classification accuracy, is rarely identical for each client. Indeed, there is always a client that profits more than what they put in. This raises the question of how to incentivise the strongest clients (\textit{i.e.,} those clients who contribute a great deal to the accuracy of their collaborators, but who do not necessarily obtain a sizeable increase in accuracy for their own model) to participate.
Central to such an incentive scheme is the question of how to quantify the contribution of each client. 

\begin{figure}    
     \begin{subfigure}[b]{0.5\columnwidth}
        \centering
        \includegraphics[width=1\linewidth]{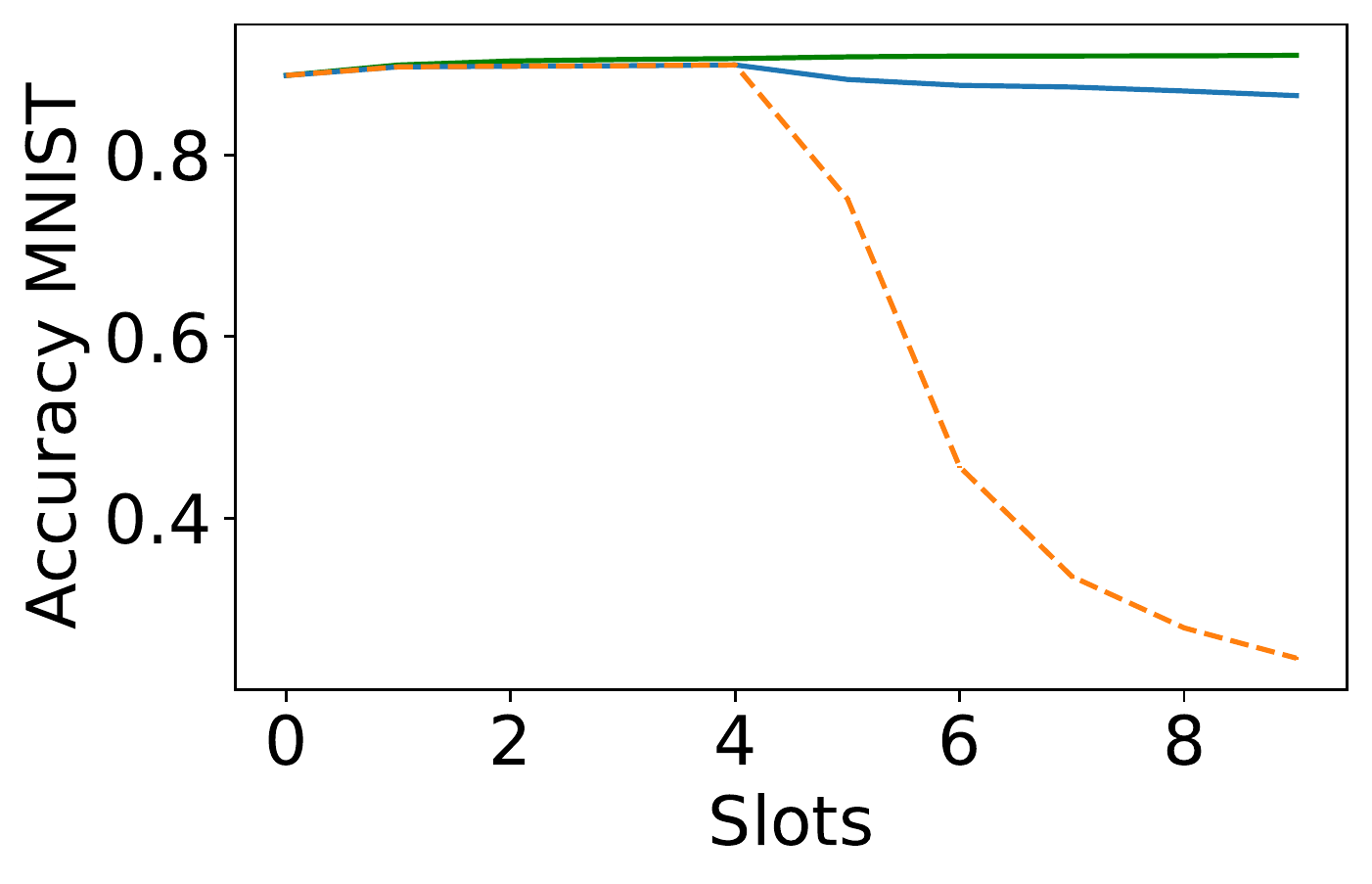}
        \caption{MNIST drops after slot 4.}
    \end{subfigure}%
    ~
    \begin{subfigure}[b]{0.5\columnwidth}
        \centering
        \includegraphics[width=1\linewidth]{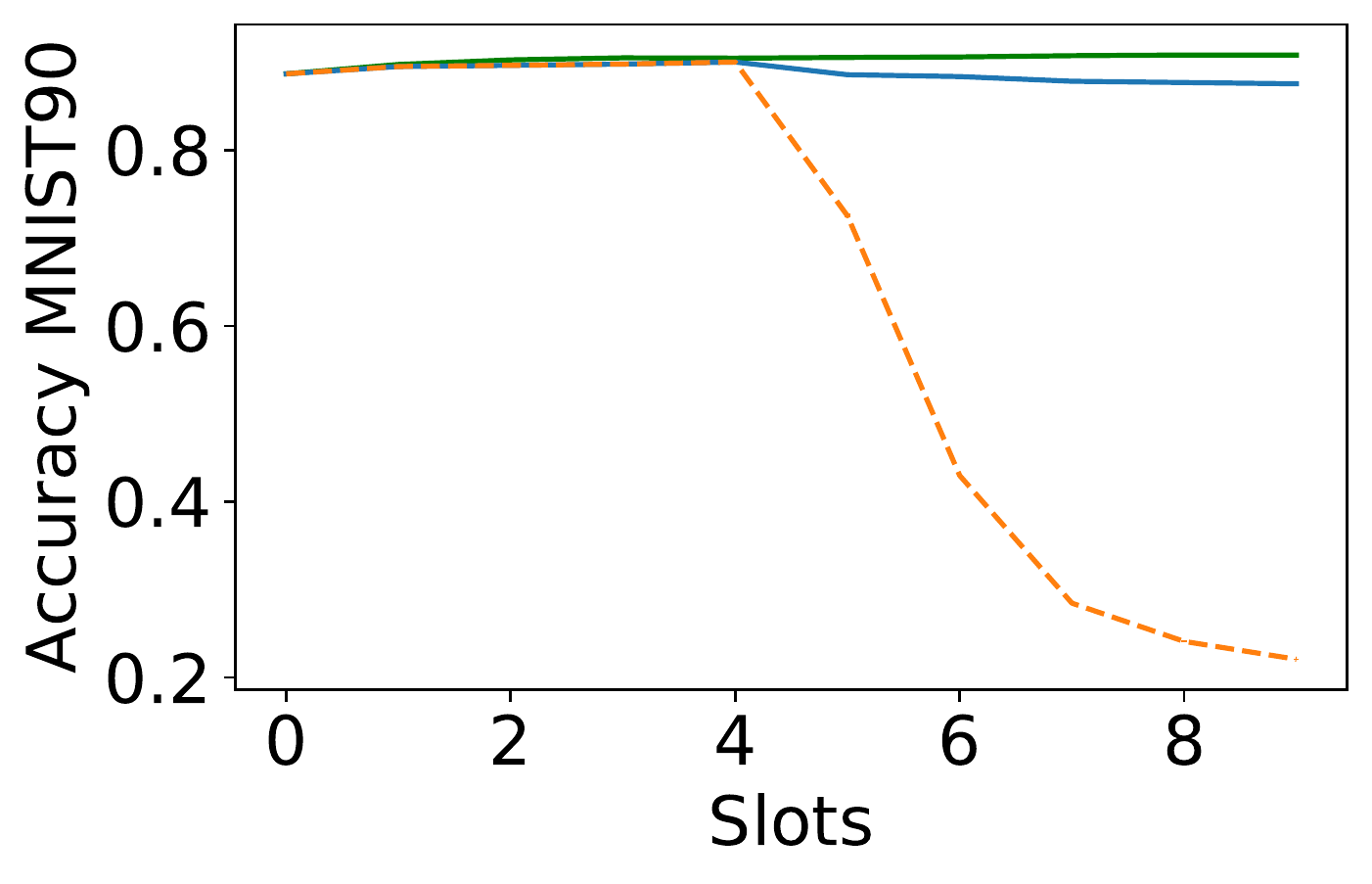}
         \caption{MNIST90 drops after slot 4.}
    \end{subfigure}%
\caption{MNIST-MNIST90 : Asynchronous training, Ideal Baseline (green), FedAvg (orange), Proposed Approach (blue). Note that standard deviation is too small for error bars to be visible on plot.}
\end{figure}

\paragraph{Disclaimer}
This paper was prepared for informational purposes by the Artificial Intelligence Research group of JPMorgan Chase \& Co.  and its affiliates (``JP Morgan''), and is not a product of the Research Department of JP Morgan. JP Morgan makes no representation and warranty whatsoever and disclaims all liability, for the completeness, accuracy or reliability of the information contained herein. This document is not intended as investment research or investment advice, or a recommendation, offer or solicitation for the purchase or sale of any security, financial instrument, financial product or service, or to be used in any way for evaluating the merits of participating in any transaction, and shall not constitute a solicitation under any jurisdiction or to any person, if such solicitation under such jurisdiction or to such person would be unlawful.

\bibliographystyle{ACM-Reference-Format}
\bibliography{references}

\appendix

\end{document}